\pdfoutput=1

\documentclass[11pt]{article}

\usepackage[final]{output}

\usepackage{times}
\usepackage{latexsym}

\usepackage{times}
\usepackage{latexsym}
\usepackage{graphicx} 
\usepackage{pdfpages} 
\usepackage{fancyhdr}
\usepackage{caption} 
\usepackage{balance}

\usepackage{placeins}
\usepackage{graphicx}
\usepackage{multirow}
\usepackage{array}
\usepackage{float}

\usepackage{booktabs} 

\usepackage{enumitem} 
\usepackage{titlesec} 

\usepackage[T1]{fontenc}

\usepackage[utf8]{inputenc}

\usepackage{microtype}

\usepackage{inconsolata}

\usepackage{graphicx}

\title{Empowering Persian LLMs for Instruction Following: \\ A Novel Dataset and Training Approach\thanks{Accepted at LoResLM @ COLING 2025}}

\author{Hojjat Mokhtarabadi{$^*$}, Ziba Zamani{$^\dagger$}, Abbas Maazallahi{$^\ddagger$}, Mohammad Hossein Manshaei{$^\S$}\\
{$^*$}Department of Electrical and Computer Engineering, Isfahan University of Technology, Iran\\
$^\dagger$Department of Computer Engineering, Shahid Bahonar University of Kerman, Iran\\
$^\ddagger$Department of Electrical and Computer Engineering, University of Tehran, Tehran, Iran\\
$^\S$Department of Computer Science, Hunter College, City University of New York, NY, USA\\
{\small h.mokhtarabadi@ec.iut.ac.ir, zibazamani@eng.uk.ac.ir, a.maazallahi@ut.ac.ir, mohammad.manshaei@hunter.cuny.edu}
}

\begin{document}

\maketitle

\begin{abstract}

Instruction-tuned large language models have demonstrated remarkable capabilities in following human instructions across various domains. However, their proficiency remains notably deficient in many low-resource languages. To address this challenge, we begin by introducing FarsInstruct: a comprehensive instruction dataset designed to enhance the instruction-following ability of large language models specifically for the Persian language—a significant yet underrepresented language globally. FarsInstruct encompasses a wide range of task types and datasets, each containing a mix of straightforward to complex manual written instructions, as well as translations from the Public Pool of Prompts, ensuring a rich linguistic and cultural representation. Furthermore, we introduce Co-CoLA, a framework designed to enhance the multi-task adaptability of LoRA-tuned models. Through extensive experimental analyses, our study showcases the effectiveness of the FarsInstruct dataset coupled with training by the Co-CoLA framework, in improving the performance of large language models within the Persian context. As of the current writing, FarsInstruct comprises 197 templates across 21 distinct datasets, and we intend to update it consistently, thus augmenting its applicability.


\end{abstract}

\section{Introduction}

The modern era of artificial intelligence is marked by numerous breakthroughs, among which is the rise of large language models (LLMs), such as GPT4~\citep{openai2024gpt4}, Llama3~\citep{dubey2024llama3herdmodels} and PaLM~\citep{chowdhery2022palm}. Instruction-tuning emerges as a vital technique in the evolution of language models, involving training a model on a wide range of tasks described through natural language instructions. This method diverges from traditional task-specific fine-tuning and adapts the model's behavior to respond to user queries with relevant and helpful answers. This technique offers a more generalized and versatile approach to model training, thus contributing significantly to the advancement of LLMs.

Despite the steady progress of instruction-tuned language models, a persistent limitation remains: their difficulty in capturing the nuanced complexities of low-resource languages. This critical challenge stems from the significant gap in the availability of high-quality instruction datasets tailored to these languages.~\citet{wang2023far} highlights this concern, demonstrating that datasets lacking sufficient multilingual diversity can cause models to lose previously learned multilingual capabilities, leading to performance degradation. Moreover, translating English-centric datasets offers only partial solutions due to several inherent limitations~\citep{naous2024having, ramesh-etal-2023-fairness, vanmassenhove-etal-2021-machine}. While efforts have been made to compile extensive multilingual instruction datasets~\citep{wang-etal-2022-super,singh2024aya,muennighoff2022crosslingual}, gaps remain in creating diverse and complex prompts for languages like Persian compared to other languages.

In this study, we propose FarsInstruct, a comprehensive human-annotated instruction dataset created from existing Persian NLP datasets. It includes a mixture of manually written instructions ranging from basic to proficient language levels, alongside translations from the Public Pool of Prompts (P3)~\citep{sanh2022multitask}, which is a collection of prompted English datasets. To ensure the diversity and representativeness of FarsInstruct, we developed 197 prompt templates derived from 21 distinct public datasets. Each prompt template comprises an input template and a target template, both of which function to extract relevant data fields from their respective datasets and reformat them into a unified structure designed for the instruction-tuning objective. For example, in the case of a Textual Entailment dataset containing the fields \textit{Premise}, \textit{Hypothesis}, and \textit{Label}, an input template might be: "Can the hypothesis be concluded from the premise? Premise: \{\textit{Premise}\}, Hypothesis: \{\textit{Hypothesis}\}", while a corresponding target template could be "The answer is: \{\textit{label}\}".

\begin{figure}[t]
\centering
\includegraphics[scale=0.05]{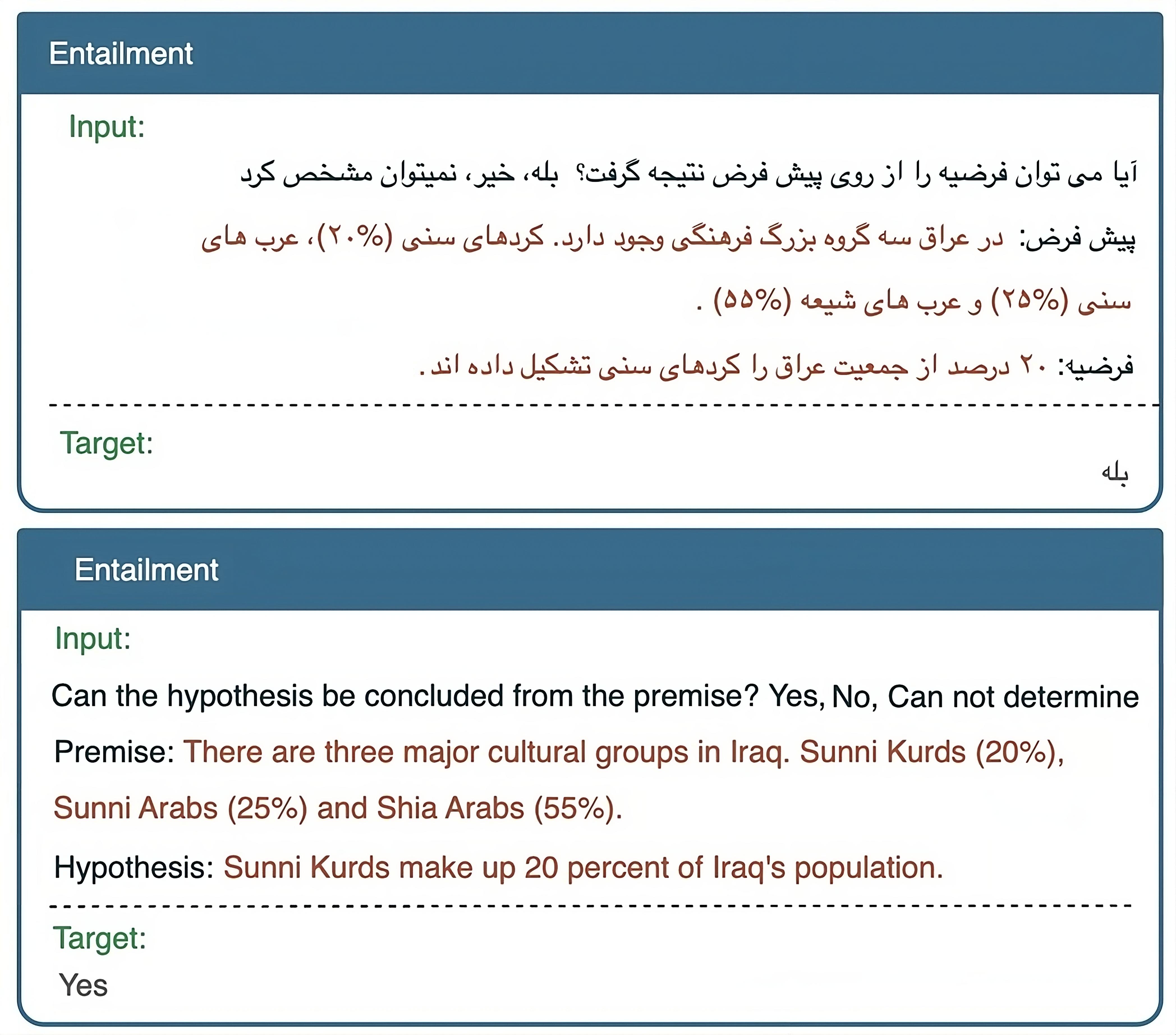}
\caption{An example of the prompts utilized in the training process. The Persian version of the prompt is employed for training purposes, while the translated English version is provided to enhance comprehension. The instruction component is highlighted in black, the data fields are marked in orange, and the target answer is indicated in gray. In Appendix~\ref{sec:appendix_prompts}, this example is shown in the PromptSource environment.}
\label{fig:sample}
\end{figure}

The collected public datasets encompass ten different task categories: Text Summarization, Textual Entailment, Text Classification, Sentiment Analysis, Word Sense Disambiguation, Query Paraphrasing, Question Answering, Reading Comprehension, Named Entity Recognition (NER), and Translation. Figure~\ref{fig:sample} depicts an instance of a prompt within our dataset after applying its respective template. A detailed overview of the FarsInsturct dataset is provided in Section~\ref{sec:farsinstruct}.

Additionally, parameter-efficient fine-tuning (PEFT) methods, such as Low-Rank Adaptation (LoRA)~\citep{hu2021lora}, not only face challenges in multi-task settings but are also prone to catastrophic forgetting~\citep{wang2023multilora,li2024mixlora,kalajdzievski2024scaling}. To address these issues, we propose Co-CoLA, a novel integration of CoLA~\citep{xia2024chain} with rehearsal training~\citep{kirkpatrick2017overcoming}. More specifically, we adopt an iterative optimization framework that merges learned low-rank matrices into the model parameters and reinitializes optimization for new LoRA modules. At each iteration, we retrain a subset of data from previously learned tasks, mixing it with the current task’s data during training. This periodic revisiting of earlier tasks ensures that the model retains performance across both old and new tasks, all while preserving computational efficiency.
Section~\ref{sec:method} presents an in-depth explanation of the Co-CoLA method.


In summary, our contributions to advancing Persian instruction understanding are threefold: (1) We present FarsInstruct, a comprehensive human-annotated instruction dataset for Persian, covering varied and representative tasks for different categories such as text summarization, named entity recognition, and translation. (2) We introduce Co-CoLA, a method that combines CoLA with rehearsal training to mitigate catastrophic forgetting in multi-task learning. (3) We release FarsInstruct as an open-source resource, with a commitment to its continued expansion to include a broader range of tasks and modalities\footnote{\url{https://huggingface.co/datasets/PNLPhub/FarsInstruct}}$^,$\footnote{\url{https://github.com/Hojjat-Mokhtarabadi/FarsInstruct}}.




\section{Related work}
\label{sec:related}
\textbf{Instruction-tuning:} 
Instruction tuning refers to the process of training language models using specific input-output pairs derived from diverse data sources. This approach enhances the ability of a pre-trained LLM to interpret and respond to a wide range of human requests expressed in natural language. Instruction datasets used for this purpose are typically created in one of three ways: (1) manually created by researchers from existing NLP datasets~\citep{wang-etal-2022-super,wei2021finetuned}, (2) synthesized by prompting proprietary models with a small, seed dataset~\citep{alpaca,wang2022self,honovich-etal-2023-unnatural}, or (3) generated entirely from scratch, involving human-written prompt-response pairs~\citep{DatabricksBlog2023DollyV2,kopf2024openassistant}. In this work, we adopt the first approach to develop FarsInstruct. Previous works such as FLAN~\citep{wei2021finetuned} and P3~\citep{sanh2022multitask} have been instrumental in advancing instruction dataset creation. FLAN encompasses over 60 NLP datasets, while P3 features more than 2,000 prompts from 177 datasets, each significantly contributing to the field. SuperNaturalInstruction~\citep{wang-etal-2022-super} further advanced the field by assembling a comprehensive benchmark featuring 1,616  expert-written NLP tasks, covering 76 unique task types, and extending support to multiple languages. xP3~\citep{muennighoff2022crosslingual} expanded on P3's groundwork by including content from 46 languages, adding new tasks like Translation and Program Synthesis that P3 had not tackled. Similarly, Aya~\citep{singh2024aya} represents a major multilingual effort, featuring an extensive dataset of 513 million instances across 114 languages. This was achieved through a global collaboration involving fluent speakers who contributed instructional content. Our dataset distinguishes itself from these collections in its depth and adaptability, especially with the inclusion of more challenging tasks in Persian, offering a high level of detail not found in many multilingual efforts. While most such projects primarily use machine translations and cover a narrow range of tasks, our dataset presents a wide array of culturally and linguistically rich tasks.

\textbf{Parameter effecient fine-tuning:} Conventional full-parameter fine-tuning becomes computationally impractical as model size and the number of downstream tasks increase. To address this challenge, recent advancements in PEFT methods advocate for training only a small subset of parameters while leaving the majority of pre-trained model parameters intact. One of the most widely utilized paradigms in PEFT is Low-Rank Adaptation (LoRA)~\citep{hu2021lora}. LoRA modifies only a small, low-rank portion of the model's weights by incorporating low-rank matrices into the model's weights during the training process. Despite the significant computational advantage of LoRA, it falls short in multi-task adaptation, Additionally,~\citet{kalajdzievski2024scaling} demonstrated that PEFT techniques, including LoRA, remain vulnerable to catastrophic forgetting, where models lose previously acquired knowledge when fine-tuned on new tasks. 
MultiLoRA~\citep{wang2023multilora} addresses the limitations of LoRA by reducing the dominance of top singular vectors, horizontally scaling LoRA modules, and altering the initialization of adaptation matrices, which leads to improved performance across multiple tasks with minimal additional parameters. MixLoRA~\citep{li2024mixlora} introduces multiple LoRA-based experts within a frozen pre-trained model using a top-k routing strategy to efficiently distribute tasks, independently configure attention layer adapters, and apply auxiliary load balance loss, significantly enhancing performance while reducing GPU memory consumption and training latency. 
Further, CoLA~\citep{xia2024chain} introduces an iterative optimization framework designed to improve the fine-tuning of LLMs by employing multiple iterations of LoRA. 
In this paper, we design Co-CoLA to address the issue of catastrophic forgetting, while ensuring an effective multi-task adaption.

\section{FarsInstruct Dataset}
\label{sec:farsinstruct}

With about 130 million\footnote{\url{https://en.wikipedia.org/wiki/Persian_language}} speakers, Persian — also referred to as Farsi in Iran — is an important language in the Middle East and Central Asia. FarsInstruct represents a project to provide a comprehensive public instruction dataset for the Persian community. As of this writing, FarsInstruct has 197 carefully designed and created prompt templates for 21 already-published public datasets and some translations from existing prompted datasets. Unlike multilingual collections focusing on common tasks such as Text Summarization and Question Answering, FarsInstruct introduces more task types, including Named Entity Recognition and Word Sense Disambiguation.  The creation procedure, statistics, task augmentation, and quality of the dataset are covered in detail in the following subsections. Additional illustrations and tables are provided in the Appendix~\ref{dataset_details},~\ref{sec:appendix_dataset_assets},~\ref{sec:appendix_prompts}. 

\subsection{Dataset Construction}


\begin{figure*}[t]
    \centering
    \includegraphics[width=0.999\textwidth]{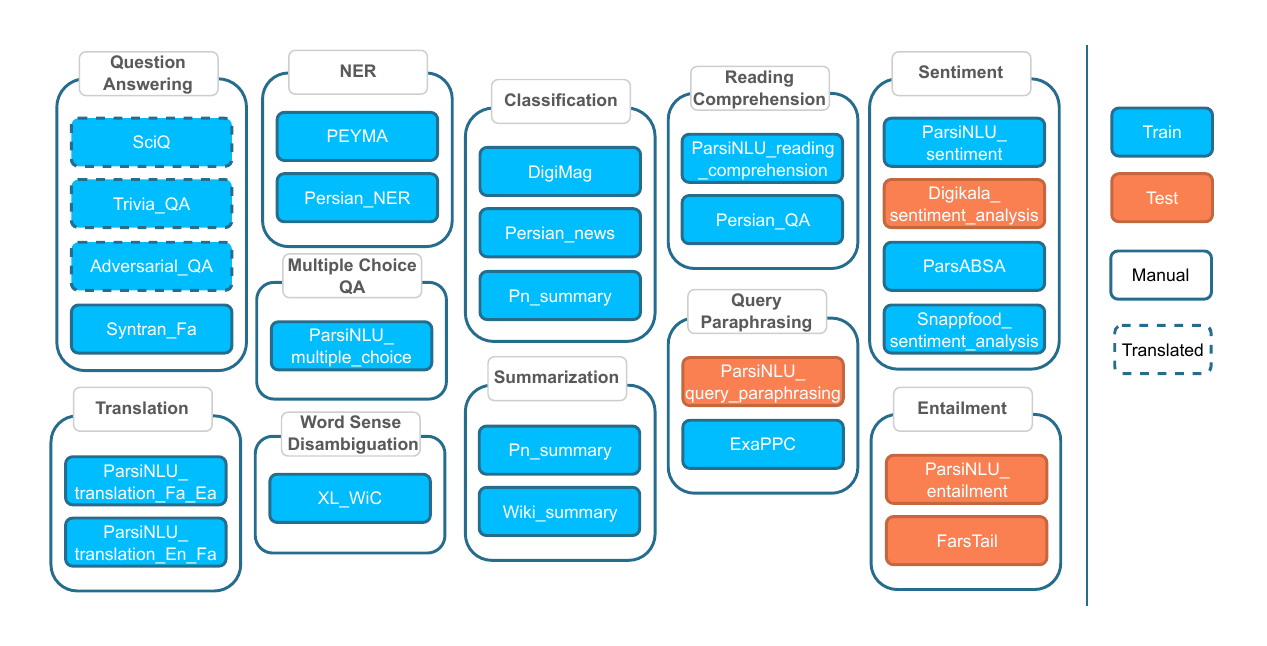}
   \caption{The detailed depiction of 11 task types utilized in our dataset. Each box within the figure lists the specific datasets associated with the respective task type. Datasets designated for training are highlighted in blue, and those reserved for testing are marked in orange. Additionally, manual datasets, which have been specifically curated and prompted by our team, are enclosed with solid borders. In contrast, datasets that have been translated from English to Persian are enclosed with dashed borders.}
    \label{fig:train_and_test_datasets}
\end{figure*}

The development of FarsInstruct entailed transforming Persian NLP datasets into their prompted format, described in plain language. This process involved a combination of manual ideation, during which our team meticulously brainstormed and refined prompt templates, along with invaluable insights from Persian language instructors. For datasets with multiple data fields, prompts were crafted to interrelate these fields, as elaborated in Section~\ref{subsec:task_augmentation}. Additionally, synonyms were employed to diversify the instructions within the prompts and reduce repetition. Each prompt template falls into one of two classes: categorization or generation. Categorization prompts guide the model in classifying text into predefined categories from dataset labels or identified through dataset analysis. In contrast, generation prompts require the model to produce full-length text, such as summarizing longer texts or answering questions based on the provided information. These instructions also include scenarios where the model needs to generate missing content from partial text inputs.

To efficiently create a large collection of prompts, we primarily utilized PromptSource~\citep{bach2022promptsource}, an open-source tool designed for creating, sharing, and managing prompts for NLP tasks. A key design choice in~\citet{bach2022promptsource} is the use of Jinja2\footnote{\url{https://jinja.palletsprojects.com/en/3.1.x/}} as a templating language, providing the flexibility crucial for crafting clear and effective prompts. Each dataset has multiple prompt templates, each of which consists of an input and a target template. These templates map raw data fields into natural language, structuring both the input and target sequences. Practically, templates allow users to mix arbitrary text with data fields. We refer to the text within the input template that guides the model’s behavior as "Instruction". Additionally, each prompt template documents essential metadata, including evaluation metrics and the language used. 

The PromptSource toolkit offers an interface for interactively writing prompts on datasets. However, the original version did not support Persian, so we modified its source code to handle Persian datasets. Our updated version is publicly available, providing the Persian community with a tool to simply create and develop prompts\footnote{\url{https://github.com/Hojjat-Mokhtarabadi/promptsource}}. Appendix \ref{sec:appendix_prompts} depicts an illustration of the PromptSource interface with an example of a Textual Entailment dataset. Moreover, since this system was originally integrated with Huggingface Datasets library~\citep{lhoest-etal-2021-datasets}, we gathered datasets from various sources and consolidated them into a unified public repository on HuggingFace. Appendix \ref{sec:appendix_prompts} provides a sample of the crafted prompt templates for different datasets.

In addition to manual templating, we have decided to translate a subset of three question-answering datasets from the P3 collection~\cite{sanh2022multitask}. This decision was made to enhance the comprehensiveness and utility of our work by providing a broader scope of data. To ensure a high-quality translation, we utilized the No Language Left Behind (NLLB)~\citep{costa2022no} machine translation model, capable of single-sentence translations between 200 languages and dialects in various scripts. We employed the largest NLLB model with 3.3B parameters to achieve the best performance. A complete list of manually templated and translated datasets is given in Figure~\ref{fig:train_and_test_datasets}.

\begin{figure}[t]
\centering
\includegraphics[scale=.68]{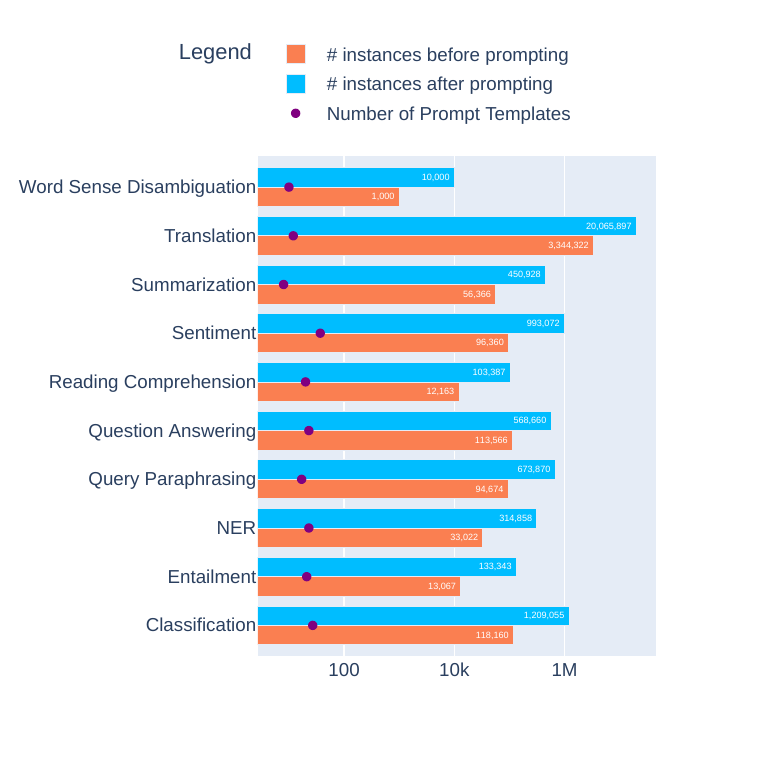}
\caption{Distribution of NLP tasks across the FarsInstruct dataset, highlighting the expanded data volumes after applying prompt templates and the number of prompts designed per task type. For each dataset, the final size is determined by multiplying the number of samples (N) by the number of prompt templates (M), resulting in a dataset size of N*M.}
\label{fig:bimodal_nlp_tasks_vertical}
\end{figure}

The final dataset is standardized through a series of preprocessing steps like deduplication and removing irrelevant elements (HTML tags, hyperlinks, emojis, and offensive language). Figure~\ref{fig:bimodal_nlp_tasks_vertical} shows the distribution of tasks across FarsInstruct, with Table~\ref{table:task_list} listing the total number of categorization and generation prompts for each task type.




\subsection{Task Augmentation and Quality Control}
\label{subsec:task_augmentation}
Instruction-tuned language models are known for their significant benefits from exposure to a broad array of tasks. In this regard, we aimed to diversify the tasks through two approaches. First, we phrased the instructions at varying language levels, ranging from basic to advanced. Second, building on best practices outlined in the FLAN Collection~\citep{longpre2023flan}, T0~\citep{sanh2022multitask}, and MetaICL~\citep{min-etal-2022-metaicl}, we enhanced task diversity by mixing and swapping different data fields within a given dataset. For instance, while a dataset may initially assess a model's ability to answer question X based on input Y, we train the model to generate question X when provided with answer Y, thereby effectively broadening the range of prompts available within a limited data pool.



\begin{table}[t]
\centering
\renewcommand{\arraystretch}{1.1} 
\setlength{\tabcolsep}{4pt} 
\begin{tabular}{p{5.5cm}| cc}
\toprule
\textbf{Task Type} & \textbf{Cat} & \textbf{Gen} \\ 
\midrule
Question Answering & 1 & 9 \\
Translation & 2 & 10 \\
NER (Named Entity Recognition) & 4 & 19 \\
Multiple Choice QA & 9 & 1 \\
Word Sense Disambiguation & 10 & 0 \\
Classification & 15 & 12 \\
Summarization & 4 & 15 \\
Reading Comprehension & 2 & 18 \\
Query Paraphrasing & 10 & 7 \\
Sentiment Analysis & 24 & 13 \\
Textual Entailment & 16 & 5 \\
\bottomrule
\end{tabular}
\caption{List of task types, along with the number of categorization and generation prompts dedicated to each task type. The expanded version of this table can be found in the Appendix~\ref{sec:appendix_dataset_assets}.}
\label{table:task_list}
\end{table}


To ensure the accuracy and cultural relevance of the instructions, we incorporated public input and expert evaluations. Feedback was gathered from 15 randomly selected individuals and three experts in Persian literature and psychology. Participants were asked to help craft instructions in various writing formats, including formal and informal styles, and to express the same instruction in different ways, then two psychology experts and one literature professor were consulted to refine the instructions. Their expertise informed revisions, ensuring that the responses were grammatically and linguistically correct and resonated with the general Persian-speaking population. Further, the datasets adopted in FarsInstruct are predominantly used for single-task fine-tuning, as their widespread use indicates higher quality.

\section{Methodology and Experimental Setup}
\label{sec:method}
To maintain our model's robustness and generalization capabilities, we integrate the CoLA framework \citep{xia2024chain} with continual learning \citep{kirkpatrick2017overcoming}. This section offers a thorough overview of the training procedure and evaluation setup.
\subsection{Training Procedure}



\begin{figure*}[t]
\centering
\includegraphics[scale=.78]{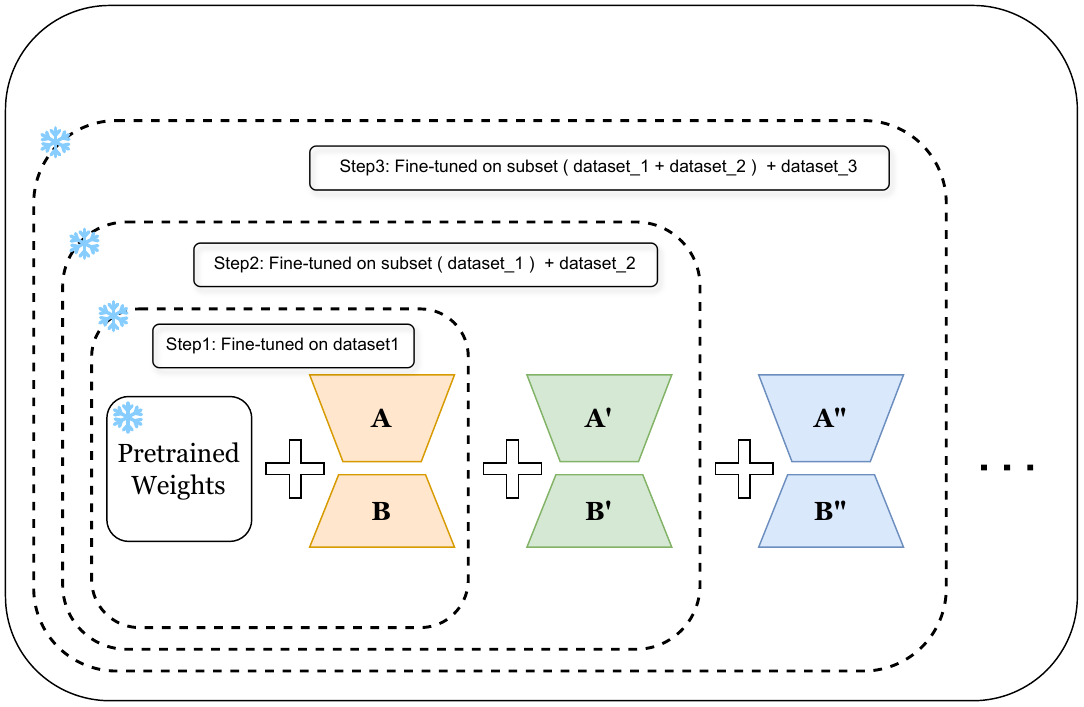}
\caption{The Continual-Chain of LoRA training procedure, containing Tuning, Merging, and Expanding. In Step 1, the pretrained language model is LoRA-tuned on dataset\_1, with the replay memory initialized as empty and merged. In Step 2, the model is expanded with a new LoRA module and further tuned on a subset of dataset\_1, determined by the rehearsal hyperparameter, alongside dataset\_2, preparing it for Step 3. This process is iteratively repeated in subsequent steps.}
\label{fig:neural_network}
\end{figure*}

Given the significant computational demands of full fine-tuning, we aim to employ LoRA for the training procedure, specifically using the FarsInstruct dataset. However, as noted in the studies by~\citep{wang2023multilora, li2024mixlora}, LoRA tends to underperform in multi-task training scenarios due to its limitations in capturing complex interactions between tasks, leading to suboptimal performance. To mitigate this challenge, Chain of LoRA (CoLA)~\citep{xia2024chain}, presents an iterative optimization framework based on the principles of the Frank-Wolfe algorithm~\citep{frank1956algorithm}. This method involves an iterative process of LoRA fine-tuning on a single task, merging the learned parameters with the base model, and reinitializing with a new LoRA module.~\citet{xia2024chain} shows this process allows the model to learn higher-rank adaptations more effectively. Another persistent challenge affecting the performance of LoRA-tuned models is catastrophic forgetting.~\citet{kalajdzievski2024scaling} analyzed this phenomenon and revealed that forgetting significantly undermines both model safety and performance on reasoning benchmarks.

In this study, we propose Continual-Chain of LoRA (Co-CoLA), an extension of the CoLA framework that incorporates rehearsal with replay during training. More specifically, rehearsal training is an approach within the continual learning framework that involves revisiting a portion of previously learned tasks while training new tasks. The core mathematical operation in LoRA involves updating the low-rank matrices \(A\) and \(B\), which are applied to modify the transformer layers of the model. The update rule can be expressed as \(W' = W + BA\) where \(W\) represents the transformer layer's original weights, and \(W'\) shows the updated weights after applying the low-rank adjustments \(A\) and \(B\). 
Essentially, Co-CoLA structures the training procedure by iterating over the following three phases:

    \textbf{Tuning:} Following the standard LoRA approach, the weights of the base model remain frozen, while only the model's LoRA parameters, represented by matrices \(A\) and \(B\) are fine-tuned. During this phase, a subset of previously trained data is replayed along with the new data. Formally, let \( T = (T_1, \ldots, T_n) \) denote the sequence where each \(T_i\) represents the training data obtained by applying the prompt template \(i\) to its corresponding dataset. The training data augmented with rehearsal is defined as: 
    \begin{equation}
        T_i^r = T_i \cup \left( \sum_{j=1}^{i-1} rT_j \right) 
    \end{equation}
        
    where r is the rehearsal hyperparameter that controls the percentage of examples sampled from previous tasks \(T_1, \ldots, T_{i-1}\).

    \textbf{Merging:} After the tuning phase, the newly updated LoRA parameters are merged with the existing model weights based on the standard method in~\citet{hu2021lora}. These merged weights are fixed and do not receive any gradient update in subsequent steps. 

    \textbf{Expanding:} The final phase involves preparing the model for subsequent training rounds by reinitializing the LoRA modules with new parameters (\(A'\) and \(B'\)). Following~\citet{hu2021lora} \(A'\) adopts Gaussian initialization and \(B'\) is initialized to zero.
    

An illustration of this iterative three-staged approach is provided in Figure~\ref{fig:neural_network}.



\subsection{Evaluation Setup}
The performance of our model is assessed across two categories of task types: those included in the training dataset ("Held in") and those introduced for the first time during evaluation ("Held out"). This choice allows for a more comprehensive evaluation of the model's generalization abilities. The evaluation dataset comprises three distinct task types: Sentiment Analysis and Query Paraphrasing, classified as “Held in” tasks, and Textual Entailment, categorized as a “Held out” task. As shown in Figure~\ref{fig:train_and_test_datasets}, the evaluation includes one dataset each for sentiment analysis and paraphrase identification, as well as two datasets specifically for entailment tasks. 

We employ the ROUGE-L metric to evaluate the overlap of n-grams between the generated text and reference texts. Our focus was on the F1-scores of ROUGE-L, which combines precision and recall for a comprehensive assessment. As shown by \citet{wang-etal-2022-super}, the rankings generated by this metric correlate strongly with accuracy for categorization templates.

\section{Results}
\label{sec:result}
To investigate the applicability of FarsInstruct, we instruction-tuned Ava---a Llama-3-based Persian LLM---using the Co-CoLA framework across a suit of tasks. The results were compared with monolingual and multilingual instruction-tuned models, using quantitative evaluations. 
For a comprehensive overview of the training configuration, please refer to the Appendix~\ref{sec:train_conf}.


\begin{table}[ht]
\centering
\begin{tabular}{l|llcc}
\toprule
\textbf{Task} & \textbf{Type} & \textbf{Model} & \textbf{ROUGE-L} \\

\midrule
\multirow{8}{*}{\rotatebox{90}{\shortstack{parsinlu query\\ paraphrasing}}} & \multirow{8}{*}{\rotatebox{90}{Held In}} & Aya-13B & 45.58 \\
 &  & PersianMind-7B & 17.07 \\
 &  & Mistral-7B & 6.89 \\
 &  & Ava-8B & 6.67 \\
 &  & Ava-LoRA-8B & 8.73 \\
 &  & CoLA-8B & 20.88 \\
 &  & Co-CoLA-8B & \textbf{45.86} \\
\cmidrule{1-4}
\multirow{8}{*}{\rotatebox{90}{\shortstack{Digikala Sentiment\\ Analysis}}} & \multirow{8}{*}{\rotatebox{90}{Held In}} & Aya-13B & 28.41 \\
 &  & PersianMind-7B & 18.19 \\
 &  & Mistral-7B & 2.46 \\
 &  & Ava-8B & 8.69 \\
 &  & Ava-LoRA-8B & 5.72 \\
 &  & Ava-LoRA-8B & 5.72 \\
 &  & CoLA-8B & 25.62 \\
 &  & Co-CoLA-8B & \textbf{40.87} \\
\cmidrule{1-4}
\multirow{8}{*}{\rotatebox{90}{\shortstack{FarsTail}}} & \multirow{8}{*}{\rotatebox{90}{Held Out}} & Aya-13B & \textbf{37.61} \\
 &  & PersianMind-7B & 17.05 \\
 &  & Mistral-7B & 5.74 \\
 &  & Ava-8B & 12.48 \\
 &  & Ava-LoRA-8B & 9.07 \\
 &  & CoLA-8B & 15.64 \\
 &  & Co-CoLA-8B & 36.35 \\
\cmidrule{1-4}
\multirow{8}{*}{\rotatebox{90}{\shortstack{Parsinlu\\ Entailment}}} & \multirow{8}{*}{\rotatebox{90}{Held Out}} & Aya-13B & 42.64 \\
 &  & PersianMind-7B & 4.45 \\
 &  & Mistral-7B & 4.93 \\
 &  & Ava-8B & 15.04 \\
 &  & Ava-LoRA-8B & 7.18 \\
 &  & CoLA-8B & 22.55 \\
 &  & Co-CoLA-8B & \textbf{55.32} \\

\bottomrule
\end{tabular}
\caption{ROUGE-L F1 Scores for Different Models across Tasks}
\label{tab:rouge_scores}
\end{table}

\subsection{Quantitative Evaluation}


\begin{figure*}[h!]
\centering
\includegraphics[scale=0.67]{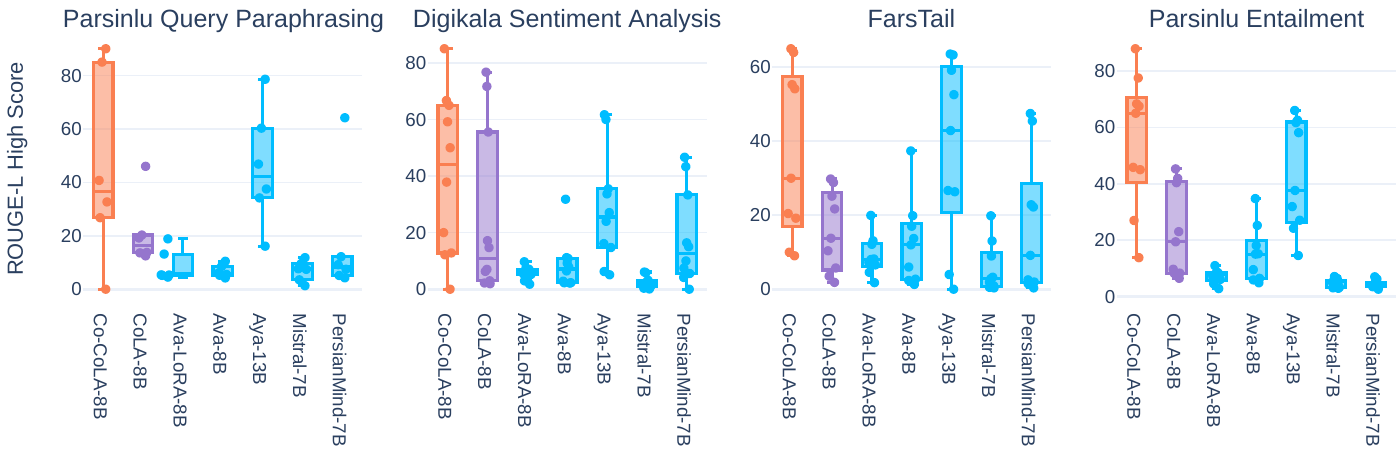}
\caption{Comparative performance of different models on Persian language tasks using the ROUGE-L metric. The bar chart depicts the superior performance of Co-CoLA across multiple tasks, particularly excelling in the ParsiNLU Entailment task.}
\label{fig:rouge_performace}
\end{figure*}

We evaluate our model against several existing models fine-tuned on instruction data. Specifically, PersianMind~\citep{huggingface_persianmind} is a Llama-2 7B-based model, trained in 3 phases on different Persian datasets. Though its training data is unavailable, Ava~\citep{huggingface_ava_llama} is a newly introduced model, fine-tuned on the Llama-3 8B model for Persian tasks. Aya~\citep{huggingface_aya_101} is a 13B encoder-decoder model trained on a subset of 25 million samples from the Aya dataset and Mistral-7B~\citep{huggingface_mistral_7b_instruct} is a decoder-only model trained on publicly available prompted datasets.

Table~\ref{tab:rouge_scores} summarizes the comparative performance of various models, including our proposed method, Co-CoLA, across several NLP Datasets: ParsiNLU Query Paraphrasing, Digikala Sentiment Analysis, FarsTail, and ParsiNLU Entailment. These models are evaluated using ROUGE-L F1 scores. As illustrated in 
Table~\ref{tab:rouge_scores}, Co-CoLA performs comparably well to the Aya model, despite having fewer parameters and being trained on less instruction data and significantly outperforms all other models, indicating the effectiveness of Co-CoLA. The factors contributing to this performance gap are further discussed in Section~\ref{sec:discussion}. Moreover, the scores of Ava-LoRA, reflecting the performance of raw LoRA fine-tuning of Ava on FarsInstruct and naive CoLA are inferior to those achieved with Co-CoLA training, highlighting the effectiveness of our method.

\section{Discussion}
\label{sec:discussion}

Figure~\ref{fig:rouge_performace} provides a detailed breakdown of the overall performance reported in Table~\ref{tab:rouge_scores}. Each dot in the plot represents the ROUGE-L F1 score of the given model on the selected template. As clearly illustrated, other Persian instruction-tuned models fail to achieve a high ROUGE-L F1 score. One significant factor contributing to this disparity is the low precision score. The F1 score combines precision and recall and serves as a comprehensive metric for evaluation. Precision measures the proportion of the longest common subsequence (LCS) in the candidate text that matches the reference text, while recall measures the proportion of the LCS in the reference text that is present in the candidate text. Although these models achieve acceptable recall scores, they fall short in precision, a critical metric for categorization templates. In contrast, Aya demonstrates proficiency in handling both generation and categorization templates within the Persian context. Compared to Aya, Co-CoLA enhances the model's ability to manage both categorization and generation tasks effectively while being less computationally expensive. Despite the limited success of continual learning frameworks, the study by~\citet{scialom2022fine} demonstrated that continual training of language models, such as T0~\citep{sanh2022multitask} with rehearsal, can effectively help them in comprehending new instructions via instruction composition. Our results confirm this finding within the Chain-of-LoRA framework, resulting in better generalization and improved performance on new tasks.


\section{Conclusion}
\label{sec:conclusion}

This study aims to address the limitations in instruction-following capabilities of language models for Persian, an important but underrepresented language, by introducing a novel instruction dataset and a training approach specifically designed to enhance the instruction comprehension of large language models. FarsInstruct presents a carefully curated dataset that combines human-annotated instruction data with translations from English-centric instruction datasets, featuring tasks in different forms and from varying levels of difficulty. Further, Co-CoLA leverages the strengths of CoLA with rehearsal training to mitigate catastrophic forgetting and improve multi-task adaptation, through its iterative optimization framework. Our results demonstrate that this allows for sustained model performance over diverse tasks while optimizing computational resources. We hope our dataset fills the critical gap and serves as a valuable resource to the multilingual NLP community.

\section{Limitations}
\label{sec:limit}
This section delineates the principal limitations of our study, which, while providing substantial contributions to Persian NLP, presents certain challenges. Addressing these challenges in future developments could enhance its utility and applicability in broader linguistic contexts.

\textbf{Data Diversity and Representation:} Although FarsInstruct broadens the corpus of Persian language resources, it does not fully capture the rich tapestry of dialects and sociolects that characterize the Persian-speaking world. Also, the collected templates are generally biased towards short responses, which might affect the overall performance of the model.

\textbf{Complexity of Instructions:} The dataset prompts vary in complexity but still may not sufficiently challenge or train models to handle the types of complex instructions encountered in everyday human interactions. Real-world applications often demand a high level of interpretative depth and context awareness—qualities that current models may struggle with when trained on existing datasets. Future versions of FarsInstruct could benefit from integrating prompts that require higher-order cognitive processing, such as irony, metaphor understanding, and techniques that involve prompting the model to break down complex tasks into intermediate steps, mimicking human reasoning processes~\citep{wei2022chain}.

    

\textbf{Dependency on External Datasets:} The effectiveness of the FarsInstruct dataset is contingent upon the quality and variety of the external datasets. This dependency creates vulnerability, as biases or errors in source datasets may be passed to FarsInstruct. A rigorous process for source data, coupled with efforts to develop original, high-quality training materials, could diminish reliance on external datasets and enhance the overall integrity of the dataset.

\textbf{Evaluation Metrics:} The metrics currently used to evaluate models trained on FarsInstruct may not fully capture the nuanced and multifaceted aspects of language comprehension and generation. Furthermore, for certain tasks such as rewriting, ROUGE-L may not serve as an adequate measure of quality. 

\textbf{Performance Stability:} While Co-CoLA has demonstrated effectiveness in terms of computational efficiency and consistent performance across all tasks it learned, mitigating catastrophic forgetting, we observe that its overall performance is heavily dependent on the model's performance at each tuning iteration. We leave potential solutions to this problem to future work.


\bibliography{colm2024_conference}

\clearpage
\onecolumn

\section*{Appendix}



\appendix

\section{Training Configuration}
\label{sec:train_conf}
\FloatBarrier 
All implementations were carried out using PyTorch~\cite{paszke2019pytorchimperativestylehighperformance}, Transformers~\citep{wolf-etal-2020-transformers} and Accelerate~\citep{accelerate} libraries. For efficient training, we randomly selected 25 prompt templates and applied them to their corresponding datasets. Consequently, for instance, a dataset with two selected templates would be upsampled to twice its original size. To generate the training data for each iteration, we sampled up to 10,000 instances from the dataset based on the selected template, with the rehearsal hyperparameter of Co-CoLA set to 0.01. Following the established practices, we used Paged-AdamW as the base optimizer and trained for a total of four epochs in each tuning phase. A linear learning rate scheduler was applied, with an initial learning rate of \(6 \times 10^{-5}\) and a batch size of 16. For implementing LoRA, we utilized the PEFT~\citep{peft} library for convenience.

\section{Datasets Details}
\label{dataset_details}
\begin{itemize}
    \item \textbf{Digikala Sentiment Analysis}~\citep{Tehranipour_2019}: A collection of Digikala product reviews labeled by customer star ratings. It categorizes sentiment into five labels (e.g., buy, not buy, neutral).
    
    \item \textbf{Snappfood Sentiment Analysis}~\citep{Tehranipour_2022}: A dataset of 70,000 user reviews from Snappfood, an online food delivery service. It contains equal numbers of positive and negative reviews (35,000 each), supporting effective sentiment analysis.
    
    \item \textbf{ParsiNLU}~\citep{khashabi-etal-2021-parsinlu}: A comprehensive suite for Persian NLP tasks, covering reading comprehension, multiple-choice question-answering, sentiment analysis, textual entailment, question-answering, and machine translation. These datasets are collected in a multitude of ways, often involving manual annotations by native speakers. This results in over 14.5k new instances across 6 distinct NLU tasks, serving as a key Persian NLP benchmark.
   
    \item \textbf{ExaPPC}~\citep{9786243}: A paraphrase corpus with 2.3 million Persian sentence pairs labeled as paraphrase or non-paraphrase. It includes both formal and colloquial sentences, making it ideal for models like BERT.
    
    \item \textbf{FarsTail}~\citep{amirkhani2023farstail}: A Persian textual entailment dataset with 10,367 samples, classifying premise-hypothesis pairs into entailment, contradiction, or neutral, essential for natural language inference in Persian.
    
    \item \textbf{Pars-ABSA}~\citep{DBLP:journals/corr/abs-1908-01815}: A dataset for aspect-based sentiment analysis in Persian, with 5,114 positive, 3,061 negative, and 1,827 neutral data points. It is useful for studying fine-grained sentiment in reviews.
    
    \item \textbf{WikiSummary}~\cite{Bert2BertWikiSummaryPersian}: A summarization dataset with 45,654 entries derived from Persian Wikipedia articles, paired with highlights designed for summarization tasks with reduced article lengths.
    
    \item \textbf{Pn-Summary}~\citep{pnSummary}: The Pn-Summary dataset contains 93,207 news articles from six news agencies, each paired with a human-generated summary. The dataset was curated from an initial pool of 200,000 articles, covering various categories. 
    
    \item \textbf{PersianQA}~\citep{PersianQA}: PersianQA is a reading comprehension dataset with over 9,000 entries sourced from Persian Wikipedia, including both answerable and unanswerable questions. It supports the development of models that can recognize unanswerable queries, similar to SQuAD 2.0. 
  
    \item \textbf{PersianNews}~\citep{ParsBERT}: This dataset consists of 16,438 news articles from online Persian news agencies, categorized into eight classes such as Economic, International, Political, Science \& Technology, and Sport. 
    
    \item \textbf{DigiMag}~\citep{ParsBERT}: DigiMag contains 8,515 articles from the Digikala Online Magazine, divided into seven categories including Video Games, Shopping Guide, Health \& Beauty, and Art \& Cinema. 
    
    \item \textbf{PEYMA}~\citep{shahshahani2018peyma}: The PEYMA dataset features 7,145 sentences with 302,530 tokens, 41,148 of which are annotated with seven entity classes, including Organization, Money, Location, Date, and Person. 
    
    \item \textbf{Persian NER}~\citep{poostchi-etal-2016-personer}: This is a manually-annotated named entity recognition dataset with 250,015 tokens and 7,682 sentences. The dataset includes six named entity classes like Person, Organization, Location, and Event, in IOB format. 
    
    \item \textbf{Syntran-fa}~\citep{farsi2024syntran}: A Farsi question-answering dataset with nearly 50,000 question-answer pairs. It extends short answers into fluent, complete responses using syntactic rules and parsing methods like stanza. 
    
    \item \textbf{XL-WiC}~\citep{raganato-etal-2020-xl}: XL-WiC is a multilingual dataset for word sense disambiguation, involving binary classification of word meaning across 12 languages, including Farsi. It evaluates models on cross-lingual semantic contextualization. 
    
    \item \textbf{SciQ}~\citep{lu2022learn}: A multimodal dataset with 21,208 science questions from elementary and high school curricula. It covers various sciences, with questions annotated with images, lectures, and explanations, making it a rich resource for science QA.
    
    \item \textbf{TriviaQA}~\citep{JoshiTriviaQA2017}: A large QA dataset with 950,000 question-answer pairs from Wikipedia and web documents. It is more challenging than datasets like SQuAD due to longer contexts and non-direct text spans, including human-verified and machine-generated pairs.
    
    \item \textbf{AdversarialQA}~\citep{Bartolo_2020}: A dataset designed to test the robustness of QA models against adversarially crafted questions. It includes adversarially modified questions from SQuAD, TriviaQA, and NewsQA to challenge model reasoning and generalization.
\end{itemize}

\newpage
\section{Datasets Illustrations}
\label{sec:appendix_dataset_assets}
\FloatBarrier 

\begin{table}[h]

\renewcommand{\arraystretch}{1.2}
\centering
\begin{tabular}{l|cc}
\toprule
Dataset & Categorization & Generation\\
\midrule

DigiMag &  9 & 1 \\
Digikala\_sentiment\_analysis &  9 & 1 \\
ExaPPC &  3 & 4 \\
FarsTail & 8 & 2 \\
Pars\-ABSA &  5 & 1 \\
ParsiNLU\_entailment & 8 & 3 \\
ParsiNLU\_multiple\_choice &  9 & 1 \\
ParsiNLU\_query\_paraphrasing &  7 & 3 \\
ParsiNLU\_reading\_comprehension &  1 & 9 \\
ParsiNLU\_sentiment &  3 & 7 \\
ParsiNLU\_translation\_En\_FA &  1 & 5\\
ParsiNLU\_translation\_FA\_En &  1 & 5\\
PEYMA &  1 & 9 \\
Persian\_NER & 3 & 10 \\
Persian\_news &  3 & 3 \\
Persian\_QA &  1 & 9 \\
Pn\_summary & 3 & 8 \\
Snappfood\_sentiment\_analysis & 7 & 4 \\
Syntran\_FA &  1 & 9 \\
Wiki\_summary &  1 & 7 \\
XL\_WiC & 10 & 0 \\

\bottomrule
\end{tabular}
\caption{Detailed Overview of Datasets Utilized for Categorization and Generation Tasks. As shown in this table Categorization and Generation tasks are not equally distributed across all datasets. Some datasets, such as Digimag, are originally designed for categorization tasks. We have enhanced these datasets by incorporating generation prompts. Conversely, translation tasks, which are inherently generative, have been augmented with categorization prompts. This dual-purpose approach enriches the datasets, facilitating both categorization and generation tasks and providing a more versatile training and testing framework.
This table provides insight into the distribution and specialization of prompts across different datasets, highlighting the balance and focus within the training and testing framework.}
\label{table:datasets_list}
\end{table}

\newpage


\begin{figure*}[h!]
\centering
\includegraphics[width=\textwidth]{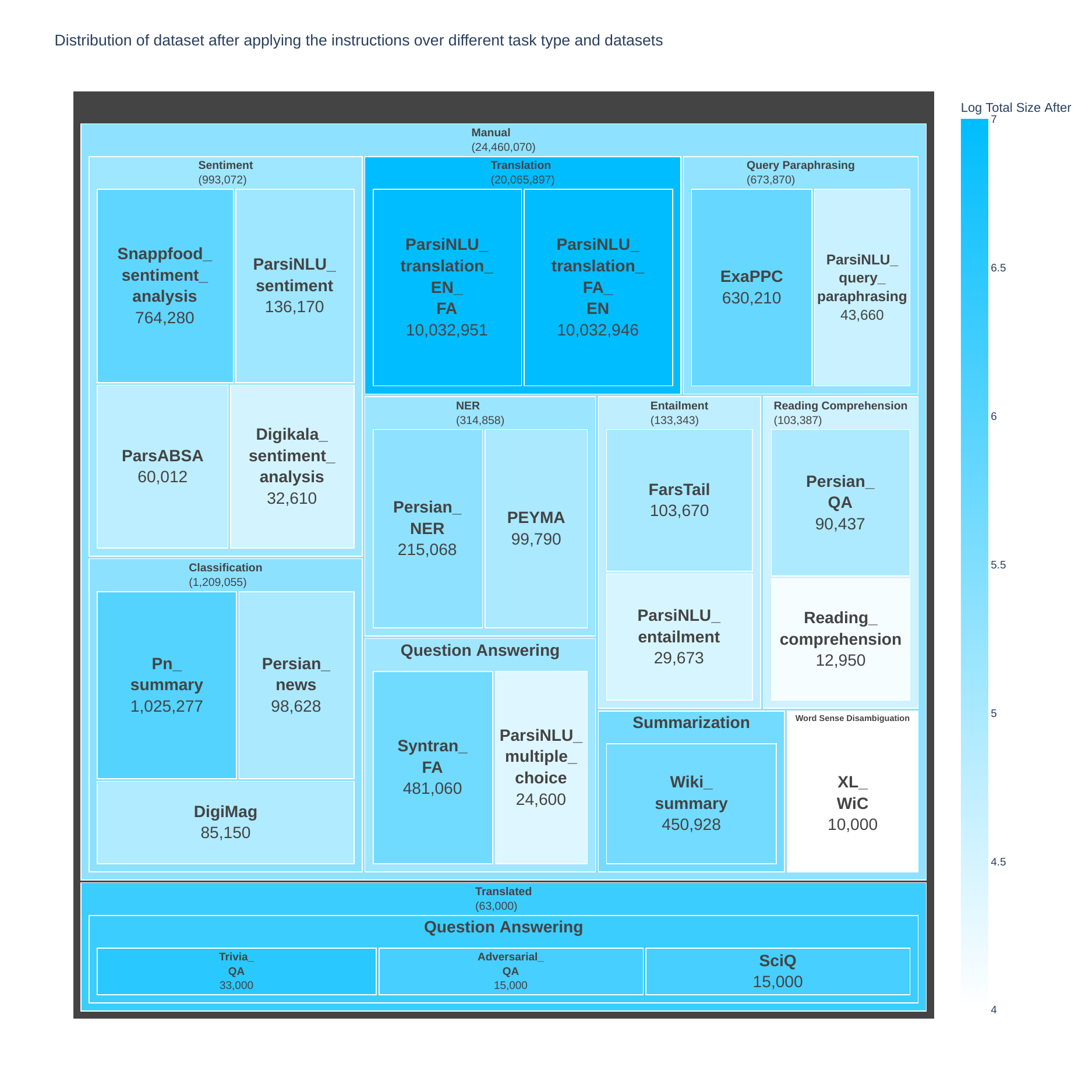}
\caption{A treemap visualization that organizes datasets by task type, post-instruction application size, and data category (training vs. testing). Each primary rectangle represents a distinct task type within the natural language processing field, encompassing areas such as Question Answering, Classification, Translation, and more. Within these primary rectangles, smaller sub-rectangles represent individual datasets. The area of each sub-rectangle is scaled to the logarithm of the size of the dataset to accommodate the broad variance in dataset sizes, ensuring a more balanced visual representation that allows for the inclusion of both large and small datasets on the same scale.}
\label{fig:treemap}
\end{figure*}

\newpage



\onecolumn

\section{Prompts}
\addcontentsline{toc}{section}{C. Prompts}
\label{sec:appendix_prompts}


\begin{figure*}[h]
    \centering
    \includegraphics[width=0.95\textwidth]{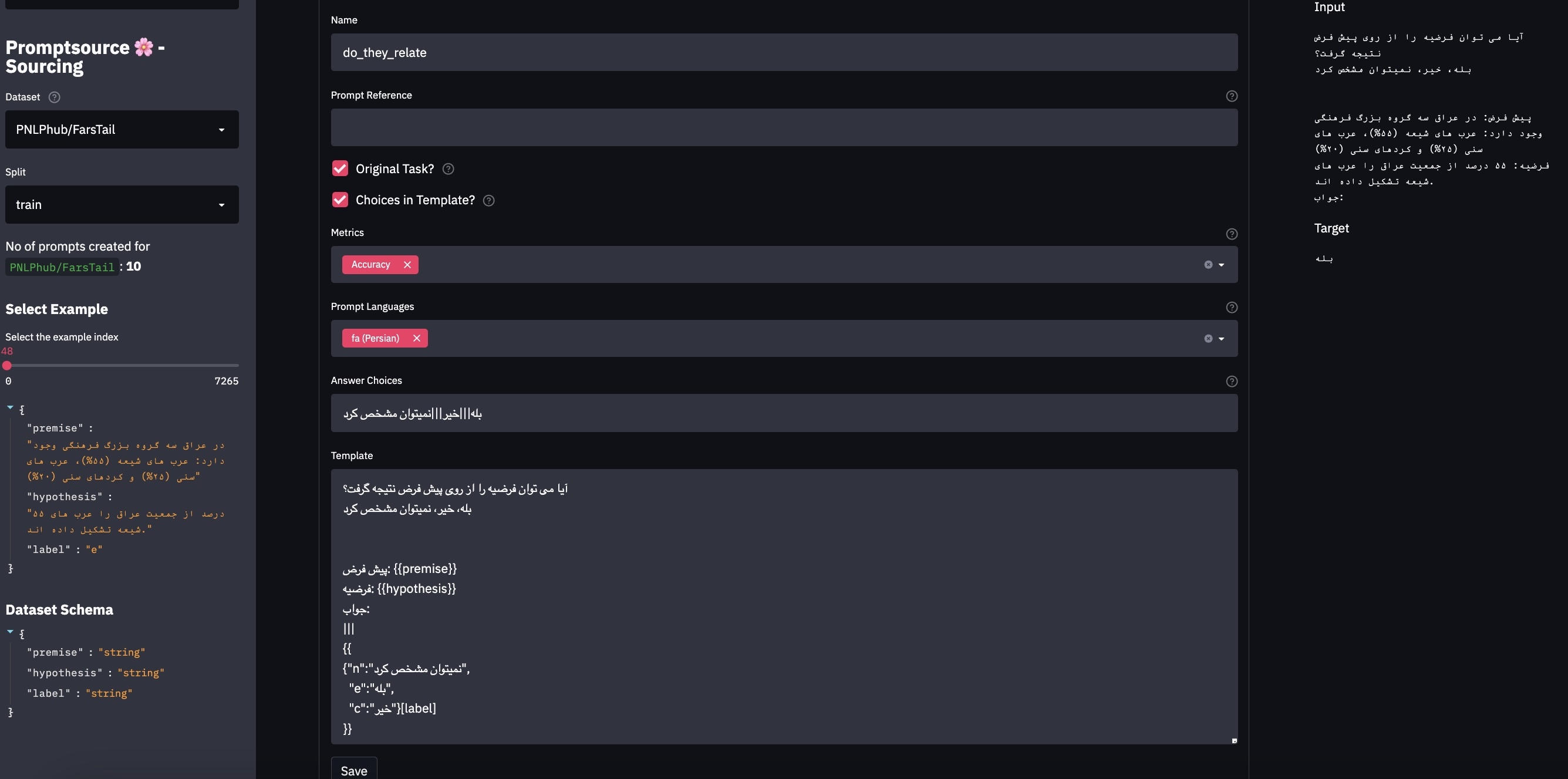}
    
    \caption{An example shown in the PromptSource environment. PromptSource is an advanced toolkit designed for creating, sharing, and utilizing natural language prompts. Prompt templates function as mappings that convert examples from datasets into natural language inputs and corresponding target outputs. In PromptSource, we develop input templates, target templates, and choice templates. Inputs typically consist of questions or instructions, while the output code specifies the expected answer or result. For categorization tasks, the choice template includes predefined options for answering questions, while generation tasks do not require this template. In this picture, The "Metrics" box is set to measure Accuracy for categorization tasks, and the "Prompt Language" used is Farsi (Persian). "Answer choices" are provided within the template, which comprises an instruction followed by data fields. The premise and hypothesis are selected from the "Data Schema" on the left side of the interface. The $|||$ symbol separates instructions from outputs, and the output employs Jinja code for conditional logic: if the label is c, it outputs 
     (no); if the label is e, it outputs 
     (yes); and if the label is n, it outputs 
     (cannot determine). 
    }
    \label{fig:promptsource_ui}
 \end{figure*}
 
\fancypagestyle{pdfinclude}{
    \fancyhf{} 
    \fancyhead{} 
    \fancyfoot{} 
    \renewcommand{\headrulewidth}{0pt}
    \renewcommand{\footrulewidth}{0pt}
}
\fancypagestyle{empty}{
    \fancyhf{} 
    \renewcommand{\headrulewidth}{0pt}
    \renewcommand{\footrulewidth}{0pt}
}


\section*{Supplementary material}

Dataset: persiannlp/parsinlu_entailment
1. GPT3_Style
Input Template:

> انتخاب کن که جمله اول و جمله دوم نسبت به هم چه نوع ارتباطی دارند؟ ارتباط منطقی وجود ندارد، مرتبط، نامرتبط
>
> جمله اول: {{sent1}}
> جمله دوم: {{sent2}}
>
> جواب:

Target Template:

> {{ {"c": "ارتباط منطقی وجود ندارد", "e": "مرتبط", "n": "نامرتبط"} [label] }}

Answer Choices Template:

> مرتبط|||نامرتبط|||ارتباط منطقی وجود ندارد

2. based_on_the_previous_passage
Input Template:

> با توجه به متن داده شده آیا میتوان عبارت را نتیجه گرفت؟
> - بله
> - خیر
> - شاید
>
> متن: {{sent1}}
> عبارت: {{sent2}}
>
> جواب :

Target Template:

> {{ {"c": "شاید", "e": "بله", "n": "خیر"} [label] }}

Answer Choices Template:

> بله|||خیر|||شاید

3. can_you_infer

Input Template:

> تصور کن که عبارت اول داده شده است. آیا براساس آن میتوان عبارت دوم را استنتاج کرد؟ از بین گزینه های داده شده انتخاب کن
> - اره
> - نه
> - شاید
>
> عبارت اول: {{sent1}}
> عبارت دوم: {{sent2}}
> جواب:

Target Template:

> {{ {"e": "اره", "c": "نه", "n": "شاید"} [label] }}

Answer Choices Template:

> اره|||نه|||شاید

4. claim_relation
Input Template:

> رابطه ی بین دو ادعای داده شده را تعیین کن: (مرتبط هست، نامشخص، مرتبط نیست)
>
> ادعای اول: {{sent1}}
> ادعای دوم: {{sent2}}
> جواب:

Target Template:

> {{ {"e": "مرتبط هست", "c": "مرتبط نیست", "n": "نامشخص"} [label] }}

Answer Choices Template:

> نامشخص|||مرتبط هست|||مرتبط نیست

5. classify
Input Template:

> نوع ارتباط این دو عبارت را در یکی از سه کلاس زیر دسته‌بندی کن
> - کلاس دلالت: با توجه به عبارت مقدم، عبارت تالی درست می‌باشد
> - کلاس تضاد: با توجه به عبارت مقدم، عبارت تالی غلط می‌باشد
> - کلاس خنثی: با توجه به عبارت مقدم، نمی‌توان درباره‌ی درست یا غلط بودن تالی نظر قطعی داد
>
> عبارت مقدم: {{sent1}}
> عبارت تالی: {{sent2}}
> جواب:

Target Template:

> {{ {"n":"کلاس خنثی", "c":"کلاس تضاد", "e":"کلاس دلالت"} [label] }}

Answer Choices Template:

> کلاس خنثی|||کلاس دلالت|||کلاس تضاد

## 6. comparison
Input Template:

> با مقایسه بین پیش گزاره اول (فرض مقدماتی) و پیش گزاره دوم (پیش گزاره) چه نتیجه ای میگیرید؟
>
> پیش گزاره اول: {{sent1}}
> پیش گزاره دوم: {{sent2}}
> نتیجه:

Target Template:

> {{ {"n":"نامعلوم", "e":"هر دو پیش گزاره مشابه هستند", "c":"پیش گزاره ها متفاوت هستند"} [label] }}

## 7. classify
Input Template:

> سطح اطمینان خود را در شباهت عبارات ارائه شده بیان کنید
> - نامطمئن
> - اطمینان پایین
> - اطمینان بالا
>
> عبارت اول: {{sent1}}
> عبارت دوم: {{sent2}}
> جواب:

Target Template:

> {{ {"n": "نامطمئن", "c": "اطمینان پایین", "e": "اطمینان بالا"} [label] }}

Answer Choices Template:

> نامطمئن|||اطمینان پایین|||اطمینان بالا

## 8. does_this_imply
Input Template:

> آیا متن دوم میتواند مفهوم متن اول باشد؟ از بین گزینه های روبرو انتخاب کن
> - بله
> - خیر
> - شاید
>
> متن اول: {{sent1}}
> متن دوم: {{sent2}}
> جواب:

Target Template:

> {{ {"n": "شاید", "e": "بله", "c": "خیر"} [label] }}

Answer Choices Template:

> بله|||خیر|||شاید

## 9. evaluate
Input Template:

> دو نظریه از دو منبع اطلاعاتی مختلف بیان شده اند. ارتباط بین آنها در کدام ارزیابی قرار دارد؟
> الف) بسیار مرتبط
> ب) نامرتبط
> ج) نامشخص
>
> نظریه اول: {{sent1}}
> نظریه دوم: {{sent2}}
> جواب:

Target Template:

> {{ {"n": "ج", "c": "ب", "e": "الف"} [label] }}

Answer Choices Template:

> الف|||ب|||ج

10. gen_sent
Input Template:

> باتوجه به جمله ی زیر یک جمله بنویس به گونه ای که نوع ارتباطشان به صورت زیر باشد
>
> نوع ارتباط: {{ {"n":"نامشخص", "e":"مرتبط", "c":"نامرتبط"}[label] }}
> جمله: {{sent1}}
> جواب:

Target Template:

> {{sent2}}

---

Dataset: PNLPhubsnappfoodsentimentanalysis
1. comment
Input Template:

> با در نظر گرفتن دیدگاه کلی مشتریان نسبت به این محصول، آیا از خریدشان راضی بودند یا نه؟
>
> دیدگاه: {{comment}}
>
> جواب:

Target Template:

```
{% if label_id == 0%}
مشتری از خریدش راضی بود
{% else %}
مشتری از خریدش راضی نبود
{% endif %}
```

## 2. feelings
Input Template:

> با در نظر گرفتن کامنت خریدار، این محصول مشتری را خوشحال یا ناامید کرده است؟
>
> دیدگاه: {{comment}}
>
> جواب:

Target Template:

```
{% if label == "HAPPY"%}
این خرید مشتری را خوشحال کرده است
{% else %}
این خرید مشتری را ناامید کرده است
{% endif %}
```

## 3. gen_sentiment
Input Template:

> عبارت ارائه شده را با دقت مطالعه کن و تصمیم بگیر که محتوای آن براساس برچسب داده شده چه حسی را منتقل میکند؟
>
> برچسب: {{label}}
> عبارت: {{comment}}
> احساس:

Target Template:

```
{% if label == "SAD"%}
ناراحت
{% else %}
خوشحال
{% endif %}
```

4. is_it_neg
Input Template:

> آیا محتوای داده شده حس منفی یا بد را به خواننده منتقل میکند؟ ارزیابی باید دقیق و براساس نحوه بیان متن باشد
>
> متن: {{comment}}
> جواب:

Target Template:

```
{% if label_id == 1%}
بله
{% else %}
خیر
{% endif %}
```

5. is_it_pos
Input Template:

> آیا متن ارائه شده دارای بار احساسی مثبت است؟
>
> متن: {{comment}}
>
> جواب:

Target Template:

```
{% if label_id == 0%}
بله
{% else %}
خیر
{% endif %}
```

### 6. possibility
Input Template:

> نظر مشتری را نسبت به جنبه های مختلف کالایی که خریداری کرده، بسنجید و تصمیم بگیرید که آیا احتمال دارد که مجدد این محصول را خریداری کند؟
>
> نظر: {{comment}}
>
> جواب:

Target Template:

```
{% if label_id == 0%}
احتمال اینکه مجدد این محصول را خریداری کند زیاد است
{% else %}
احتمال اینکه مجدد این محصول را خریداری کند کم است
{% endif %}
```

### 7. rate
Input Template:

> فرم نظرسنجی از مشتری دریافت شده است و به صورت زیر میباشد. چه امتیازی به آن
> میدهید؟
> - پنج ستاره
> - یک ستاره
>
>
> فرم نظرسنجی: {{comment}}
> امتیاز:

Target Template:

> {% if label == "HAPPY"%}
> پنج ستاره
> {% else %}
> یک ستاره
> {% endif %}

Answer Choices Template:

> یک ستاره|||پنج ستاره

8. what_is_sentiment  
Input Template:

> کاربری پس از خرید یک محصول نظر زیر را در مورد آن دارد. بررسی کن که آیا او از
> خریدش خوشحال است یا ناراحت؟
>
> نظر: {{comment}}
> جواب:

Target Template:

> {{ {"SAD": "ناراحت", "HAPPY": "خوشحال"} [label] }}

Answer Choices Template:

> خوشحال|||ناراحت

## 0.1 Prompts (Translated to english)

Dataset: persiannlpparsinlu_entailment

1. GPT3_Style
Input Template:

> Choose what kind of relationship exists between the first and second sentence? No logical connection, Related, Unrelated
>
> First sentence: {{sent1}}
> Second sentence: {{sent2}}
> Answer:

Target Template:

> {{ {"c": ,"Unrelated" "e": ,"Related" "n": "No logical connection" } [label] }}

Answer Choices Template:

> Related|||Unrelated|||No logical connection

2. based_on_the_previous_passage
Input Template:

> Based on the given text, can the statement be concluded?
> - Yes
> - No
> - Maybe
>
> Text: {{sent1}}
> Statement: {{sent2}}
> Answer :

Target Template:

> {{ {"c": ,"No" "e": ,"Yes" "n": "Maybe" } [label] }}

Answer Choices Template:

> Yes|||No|||Maybe

3. can_you_infer  
Input Template:

> Imagine the first statement is given. Based on that, can the second statement be inferred? Choose from the given options
> - Yes
> - No
> - Maybe
>
>
> First Statement: {{sent1}}  
> Second Statement: {{sent2}}  
> Answer:

Target Template:

> {{ {"n": ,"Maybe" "c": ,"No" "e": "Yes" } [label] }}

Answer Choices Template:

> Yes|||No|||Maybe

4. claim_relation  
Input Template:

> Determine the relationship between the two given claims: (Related, Uncertain, Unrelated)
>
>
> First Claim: {{sent1}}  
> Second Claim: {{sent2}}  
> Answer:

Target Template:

> {{ {"n": ,"Uncertain" "e": ,"Related" "c": "Unrelated" } [label] }}

Answer Choices Template:

> Uncertain|||Related|||Unrelated

5. classify
Input Template:

> Classify the relationship between these two statements into one of the three categories below
> - Implication class: Considering the premise, the subsequent statement is correct
> - Contradiction class: Considering the premise, the subsequent statement is incorrect
> - Neutral class: Considering the premise, it's not possible to definitively state whether the subsequent statement is correct or incorrect
>
>
> Premise: {{sent1}}
> Subsequent statement: {{sent2}}
> Answer:

Target Template:

> {{ {"n": ,"Neutral class" "c": ,"Contradiction class" "e": ,"Implication class" } [label] }}

Answer Choices Template:

> Neutral class|||Implication class|||Contradiction class

6. comparison
Input Template:

> By comparing the first premise (preliminary assumption) and the second premise, what conclusion do you draw?
>
>
> First premise: {{sent1}}
> Second premise: {{sent2}}
> Result:

Target Template:

> {{ {"n": ,"Unknown" "e": ,"Both premises are similar" "c": "The premises are different" } [label] }}

7. classify  
Input Template:

> Express your confidence level in the similarity of the given statements
> - Uncertain
> - Low confidence
> - High confidence
>
>
> First statement: {{sent1}}
> Second statement: {{sent2}}
> Answer:

Target Template:

> {{ {"n": ,"Uncertain" "c": ,"Low confidence" "e": ,"High confidence" } [label] }}

Answer Choices Template:

> Uncertain|||Low confidence|||High confidence

8. does_this_imply  
Input Template:

> Can the second text be the meaning of the first text? Choose from the options
> - Yes
> - No
> - Maybe
>
>
> First text: {{sent1}}
> Second text: {{sent2}}
> Answer:

Target Template:

> {{ {"c": ,"No" "e": ,"Yes" "n": ,"Maybe" } [label] }}

Answer Choices Template:

> Yes|||No|||Maybe

9. evaluate
Input Template:

> Two theories from different information sources are stated. In which evaluation do their relationships belong?
> a) Highly related
> b) Unrelated
> c) Uncertain
>
>
> First theory: {{sent1}}
> Second theory: {{sent2}}
> Answer:

Target Template:

> {{ {"n": ,"Uncertain" "c": ,"Unrelated" "e": ,"Highly related" } [label] }}

Answer Choices Template:

> Uncertain|||Unrelated|||Highly related

10. gen_sent
Input Template:

> Considering the sentence below, write a sentence such that their relationship is as follows
>
> Relationship type: {{{"n":"Uncertain", "e":"Related", "c":"Unrelated"}[label]}}
> Sentence: {{sent1}}
> Answer:

Target Template:

> {{sent2}}

---

Dataset: PNLPhub/snappfood-sentiment-analysis
1. comment
Input Template:

> Considering the overall customer perspective towards this product, were they satisfied with their purchase?
>
> Perspective: {{comment}}
> Answer:

Target Template:

> {% if label_id == 0 %}
> The customer was satisfied with their purchase
> {% else %}
> The customer was not satisfied with their purchase
> {% endif %}

2. feelings
Input Template:

> Considering the buyer's comment, did this product make the customer happy or disappointed?
>
> Perspective: {{comment}}
> Answer:

Target Template:

> {% if label == "HAPPY"%}
> This purchase made the customer happy
> {% else %}
> This purchase disappointed the customer
> {% endif %}

3. gen sentiment
Input Template:

> Carefully read the provided statement and decide what emotion it conveys based on the given label.
>
> Label: {{label}}
> Statement: {{comment}}
> Emotion:

Target Template:

> {% if label == "SAD"%}
> Sad
> {% else %}
> Happy
> {% endif %}

4. is it neg
Input Template:

> Does the given content convey a negative or bad feeling to the reader? The evaluation should be precise and based on the way the text is expressed.
>
> Text: {{comment}}
> Answer:

Target Template:

> {% if $label_id == 1$%}
> Yes
> {% else %}
> No
> {% endif %}

5. is it pos
Input Template:

> Does the presented text have a positive emotional charge?
>
> Text: {{comment}}
> Answer:

Target Template:

```
{% if label_id == 0%}
Yes
{% else %}
No
{% endif %}
```

6. possibility
Input Template:

> Assess the customer's opinion on various aspects of the product they purchased and decide whether there is a likelihood of repurchasing it?
>
> Opinion: {{comment}}
> Answer:

Target Template:

```
{% if label_id == 0%}
The likelihood of repurchasing this product is high
{% else %}
The likelihood of repurchasing this product is low
{% endif %}
```

7. rate
Input Template:

> A customer feedback form has been received as follows. What rating would you give it?
> - Five stars
> - One star
>
> Feedback form: {{comment}}
> Rating:

Target Template:

```
{% if label == "HAPPY"%}
Five stars
{% else %}
One star
{% endif %}
```

Answer Choices Template:

```
One star|||Five stars
```

8. what is sentiment
Input Template:

```
 A user has the following opinion about a product they purchased. Determine whether they are happy or sad about their purchase.


Opinion: {{comment}}
Answer:
```

Target Template:

```
{{ {"SAD": ,"Sad" "HAPPY": "Happy" [label] }}
```

Answer Choices Template:

```
Happy|||Sad
```





\end{document}